\documentclass{ieeeaccess}
\usepackage{cite}
\usepackage{amsmath,amssymb,amsfonts}
\usepackage{algorithm2e}
\usepackage{subfigure}
\usepackage{algorithmic}
\usepackage{graphicx}
\usepackage{textcomp}
\usepackage{bbm}
\usepackage{multirow}
\usepackage{cancel}
\def\BibTeX{{\rm B\kern-.05em{\sc i\kern-.025em b}\kern-.08em
    T\kern-.1667em\lower.7ex\hbox{E}\kern-.125emX}}

\usepackage[pagebackref,breaklinks,colorlinks]{hyperref}

\usepackage{xcolor}

\begin{document}
\history{Date of publication xxxx 00, 0000, date of current version xxxx 00, 0000.}
\doi{10.1109/ACCESS.2017.DOI}

\title{Offset-Guided Attention Network for Room-Level Aware Floor Plan Segmentation}
\author{\uppercase{zhangyu wang}\authorrefmark{1} and
\uppercase{ningyuan sun\authorrefmark{2}}.}
\address[1]{Avon Old Farms School, CT, USA (e-mail: wzylite@gmail.com)}
\address[2]{University of Wisconsin-Madison, WI, USA, (e-mail: nsun28@wisc.edu)}

\markboth
{Zhangyu Wang \headeretal: Offset-Guided Attention Networks for Room-Level Aware Floor Plan Segmentation}
{Zhangyu Wang \headeretal: Offset-Guided Attention Networks for Room-Level Aware Floor Plan Segmentation}


\begin{abstract}
Recognition of floor plans has been a challenging and popular task. Despite that many recent approaches have been proposed for this task, they typically fail to make the room-level unified prediction. Specifically, multiple semantic categories can be assigned in a single room, which seriously limits their visual quality and applicability.
In this paper, we propose a novel approach to recognize the floor plan layouts with a newly proposed Offset-Guided Attention mechanism to improve the semantic consistency within a room. In addition, we present a Feature Fusion Attention module that leverages the channel-wise attention to encourage the consistency of the room, wall, and door predictions, further enhancing the room-level semantic consistency. 
Experimental results manifest our approach is able to improve the room-level semantic consistency and outperforms the existing works both qualitatively and quantitatively.

\end{abstract}

\begin{keywords}
floor plan recognition, attention, offset-guided, convolutional neural network
\end{keywords}

\titlepgskip=-15pt

\maketitle

\section{Introduction}
\label{sec:introduction}
\begin{figure}
\begin{center}
\subfigure[Ground Truth]{\includegraphics[width=2.05cm, trim={0.8cm 0cm 0cm 0cm},clip]{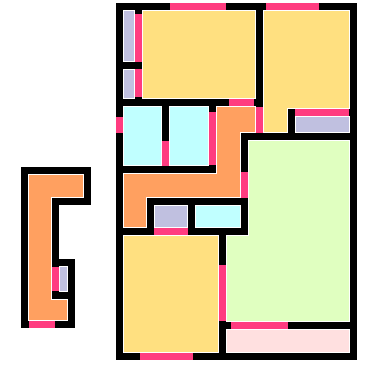}}
\subfigure[DFP]{\includegraphics[width=1.98cm, trim={0cm 0cm 0.3cm 0cm},clip]{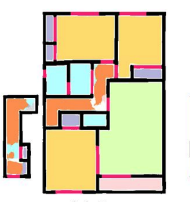}}
\subfigure[DeepLabV3+]{\includegraphics[width=2.15cm, trim={0cm 0cm 0.4cm 0cm},clip]{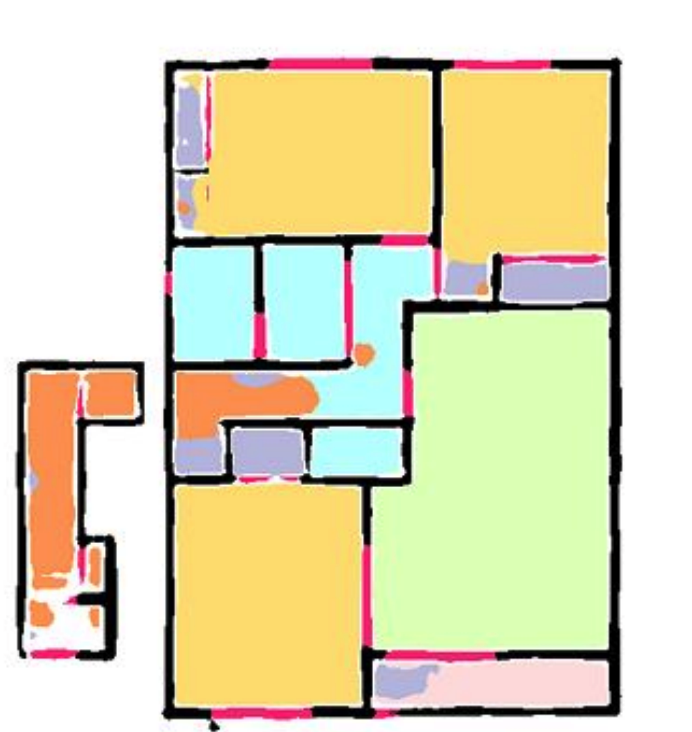}}
\subfigure[Ours]{\includegraphics[width=2cm, trim={0.85cm 0cm 0.3cm 0cm},clip]{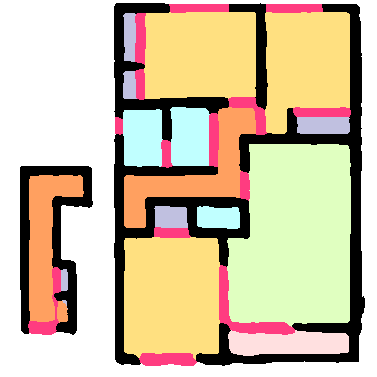}}
\caption{Existing works including Deep Floor Plan(DFP) and DeepLabV3+~\cite{b1, b20} have inconsistent room-level prediction (b,c). Our approach addresses this issue and can produce plausible results (d). }
\label{fig:fig1}
\end{center}
\end{figure}

\PARstart{T}{he} recognition of floor plan elements has been a useful task that provides benefits to various applications such as indoor robotics navigation\cite{b45}, computer-aided home design\cite{b1, b3}, and VR/AR\cite{b46, b47}. However, it is a challenging task beyond the general semantic segmentation problem. First, different from the common objects in natural images, rooms in the floor plan images typically share the similar rectangular shape as shown in Fig~\ref{fig:fig1}, so it is challenging to identify their semantic categories simply based on their appearances. 
Second, the relative spatial locations of rooms are crucial to  floor plan image understanding. However, traditional CNNs for semantic segmentation typically possess the translational invariance property, incapable of effectively capturing the rooms' relative spatial information in floor plan images. 
Third, the floor plan images typically lack textures, which plays a key role in CNNs' recognition as suggested by existing works like \cite{b49}. 
Therefore, directly adopting the semantic segmentation networks~\cite{b20, b21} designed for natural images to this task cannot produce satisfactory results due to the above issues, and quantitative proofs are provided in our experiments.




Recently, many approaches\cite{b1,b2,b14} have been proposed to address the floor plan recognition task through convolutional neural networks and provided promising results. However, their models typically lack the capability of room-level understanding. To be more specific, they predict the semantic label of each pixel individually and may assign multiple semantic categories in one single room as shown in Fig~\ref{fig:fig1} (b, c), which severely limits their applicability in practical applications. The above issue is potentially induced by the following causes. First, the rooms vary from each other significantly in terms of size; yet, their approaches cannot effectively capture both the global and local information to handle the large and tiny sized rooms respectively. Second, they typically predict the room and boundary categories (wall, door) parallelly, thus the predictions on these categories may be inconsistent with each other. A simple approach to address this issue is to leverage flood fill, i.e., assign all the pixels in a room to be the same category according to majority voting. However, this approach still cannot work well due to the reasons that will be discussed in Section~\ref{sec:related}.


In this paper, we propose a novel approach to address the above issues in existing approaches to improve the prediction consistency within a single room and the overall floor plan segmentation performance. First, we design an Offset-Guided Attention network to encourage consistency within each room. To do this, we utilize a simple prior that ``all the pixels of the same room share the same semantic category'', which always holds true in all the existing datasets as far as we know, but is ignored by the current state of the arts. 
Our network explicitly learns the correlation of pixels within each room with our newly proposed offset prediction module, and the learned correlation is further utilized to refine the output and improve the consistency of predictions within each room. 
In addition, our Offset-Guided Attention module helps the network to learn long-range context information and increase the effective receptive field of the network; hence, our approach can handle large and long rooms and have consistent semantic predictions in them (see Fig~\ref{fig:fig1} (d)). 


Besides, we propose a Feature Fusion Attention module to further improve the consistency between room and boundary predictions. Existing works~\cite{b1} propose to predict the room and boundary categories in individual branches to improve the results, which has been proven to be effective. However, this strategy has a side-effect that the prediction of the two branches may be inconsistent with each other. As shown in Fig~\ref{fig:fig1} (b), the blue room is dilated into the orange room while the wall prediction does not recognize any boundaries between blue and orange, which shows that the prediction of the boundary categories is not aligned with the room. Similarly, there are also cases where a boundary is successfully recognized by the boundary branch, but not recognized by the room branch. This is another cause of the semantic inconsistency within a single room. To address this issue, we present a Feature Fusion Attention module to combine the prediction of two branches and yield consistent predictions (Fig~\ref{fig:fig1} (d)). 


Quantitative experiments manifest that our approach outperforms the existing works on two commonly used floor plan recognition datasets\cite{b4, b13} consistently. In addition, qualitative comparisons show that our Offset-Guided Attention and Feature Fusion Attention effectively improve the room-level consistency. \textbf{Code and models will be released upon publication. }

To summarize, this paper has the following contributions to floor plan recognition :
\begin{enumerate}
\item We propose an effective and interpretable framework for floor plan segmentation beyond the existing works;
\item We propose an Offset-Guided Attention module and a channel-wise Feature Fusion Attention module to improve the room-level semantic prediction; 
\item We achieve state-of-the-art performance on the two common floor plan recognition datasets R2V~\cite{b4} and R3D~\cite{b13}. 
\end{enumerate}

\Figure[t!][scale = 0.75]{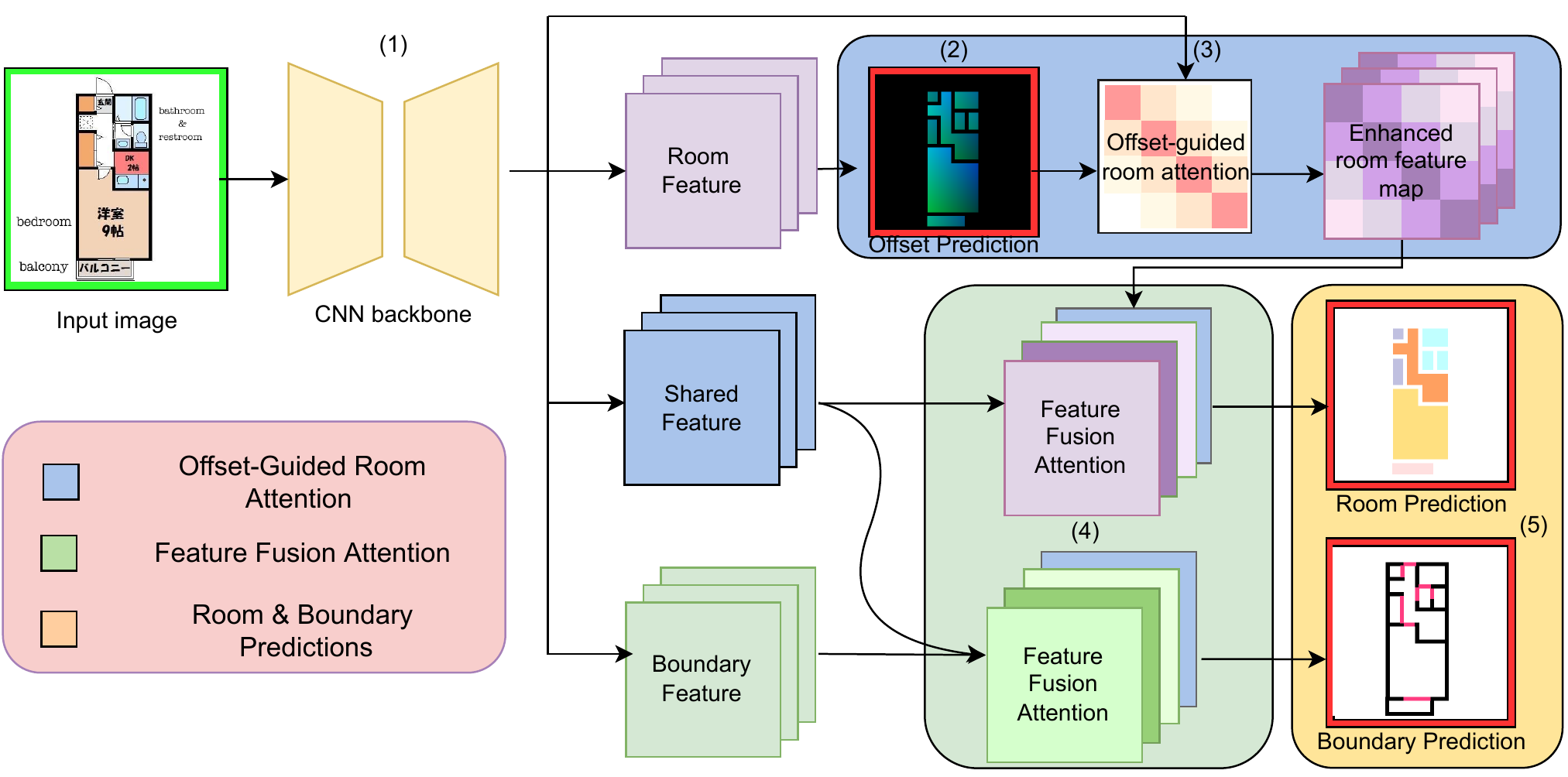}{Overview of our framework. We extract features using a CNN backbone(1); then use the feature to generate an offset prediction(2), which will later be used to determine the affinity in the Offset-Guided room attention module(3); besides, an additional feature extraction branch is used to enrich the room and boundary features by fusing them with the Feature Fusion Attention module(FFA)(4), and then produce the final boundary and room prediction(5). Note that the English labels on the ``input image'' are added for reader's reference and do not exist in the original image.\label{fig:overall}}

\section{Related Work}~\label{sec:related}

\subsection{Semantic Segmentation}
Semantic segmentation aims to recognize the pixel-wise semantic categories in the image. 
After Long et al.\cite{b17} proposes to utilize the Fully Convolutional Networks to address this issue, remarkable signs of progress are made in this field \cite{b18, b20, b21, b22, b23,b24,b25,b26,b27,b28, b29, b30, b41, b42, b43, b44}. As discussed in Section~\ref{sec:introduction}, floor plan segmentation is drastically different from the general semantic segmentation. In this work, we propose some particularly designed novel modules for floor plan segmentation and manifest their effectiveness in this challenging task. 

\subsection{Traditional Approaches on Floor Plan Recognition}
Floor plan recognition aims to derive the pixel-level prediction of a floor plan image. However, it is notably different from semantic segmentation, as discussed in Section~\ref{sec:introduction}.
This task attracts a lot of attention from both academia and industry in the past few years~\cite{b6,b7,b5,b8,b2,b1,b14,b34,b35,b36,b37}. Traditional methods \cite{b6, b7} use low-level information like graphical shapes to predict rooms and boundaries. Ahamed et al. \cite{b5} propose to decouple the boundary recognition and text recognition from the room recognition, and they introduce to extract line segments of different thicknesses -- the thicker ones are considered to be the walls and the thinner ones are the texts. In addition, another approach \cite{b8} utilizes a heuristics method for this task. However, the performance of this heuristics-based approach is far from satisfactory as the assumptions of this approach cannot cover all the circumstances. Besides, models using this heuristics-based approach need to be fine-tuned manually in a very frequent manner and many parameters need to be tuned for specific datasets.

\subsection{Room-level consistency in Floor Plan Recognition}

Recently, a variety of approaches\cite{b1,b2,b14} are proposed to address the floor plan recognition task using convolutional neural networks, which yielded promising results. However, as discussed in Section~\ref{sec:introduction}, they typically predict inconsistent semantic categories within one single room (see Fig~\ref{fig:fig1} (a,c,d)), which implies that their models lack of the room-level understanding.

An intuitive idea to address this issue is to leverage a post-processing method flood fill, which simply assigns all pixels within a certain boundary to the same semantic label according to the majority voting. An intuitive idea to address this issue is to leverage a post-processing method, flood fill, which assigns all pixels within a specific boundary to the same semantic label according to the majority voting. However, it cannot achieve satisfactory performance. That is because the boundary (door and wall categories) segmentation that flood fill is based upon a challenging task. Specifically, in many cases, the boundaries are predicted in an inaccurate or even wrong location. In addition, flood fill requires the boundary to be continuous and form a closed space to conduct the majority voting -- any slight failure in boundary prediction could cause a huge room-level prediction error, as shown in Fig~\ref{fig:fig1} (d). 

Some existing works try to enhance the accuracy of boundary prediction. However, flood fill still cannot work well based on their approaches due to their unsatisfactory boundary prediction performance. However, due to unsatisfactory boundary prediction performance, flood fill still cannot work well on the top of their approaches. Liu et al.\cite{b4} design a network that recognizes the junction point of the floor plan and determines the position of boundaries based on such prediction. However, the method can only recognize horizontal and vertical boundaries and cannot work well for others, which significantly limits its applicability. Besides, Zeng et al.\cite{b1} design a network with boundary-aware kernels and successfully bring up the accuracy of boundary predictions. Nevertheless, their boundary accuracy is still not high enough for flood fill, and the network has an inferior accuracy in certain types of rooms, where in many cases the majority of pixels predicted in a room are of the wrong category. In a word, even though existing works \cite{b4,b1} address on the boundary accuracy, the simple approach like flood filling still fails to improve the room prediction consistency and the overall accuracy.


\subsection{Attention Mechanism in Floor Plan Recognition}

Early works like \cite{b11} propose using self-attention mechanism to increase awareness of the correlation between different words in a sentence. Recently, some existing works adopt the attention mechanism in floor plan recognition. Zeng et al.\cite{b1} is the first to use attention in this task. They design a direction-aware kernel supported by room-boundary guided attention to encourage recognition of the spatial relationship between boundaries and rooms. With such a method, the accuracy of wall prediction is largely increased, but room prediction remains to be an issue. Recently, Lv et al.\cite{b3} further enhance the accuracy of the room and boundary prediction with the architecture to jointly detect and learn text, symbol, structural element, and scale information. However, since their approach relies on the text and symbol prediction, \cite{b3} only works on their own dataset but cannot be applied to the general floor plan datasets like R3D and R2V, since the symbols in these datasets have different styles and the texts are even in different languages. Besides, text and symbol recognition usually cause significant increase in the network complexity.

\Figure[t!][scale = 0.6]{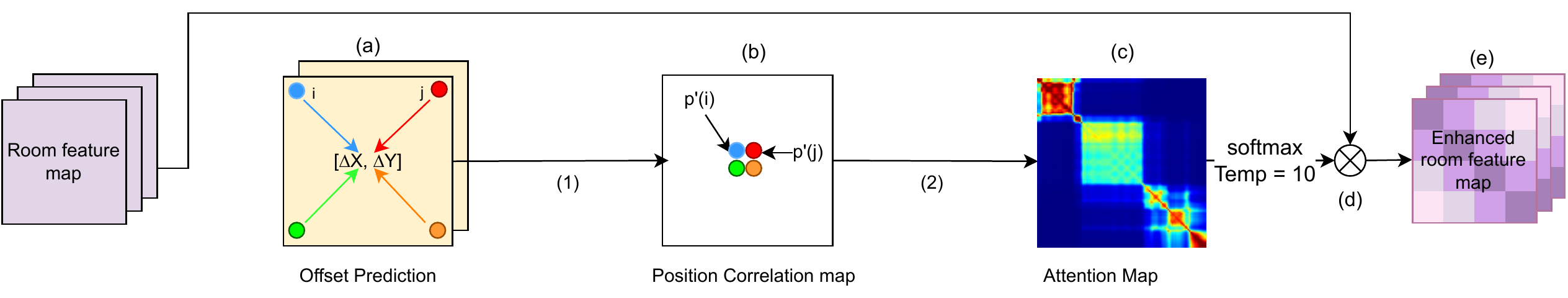}{Architecture of the Offset-Guided Room Attention module. The (1) and (2) in the image refers to Equation \ref{equ1} and Equation \ref{equ2} respectively. We use the offset prediction (a) to generate a position correlation map among pixels (b), which will then be used to derive the affinity of pixels in the attention map (c) and fused with room feature map using batch multiplication at (d). The module outputs the enhanced room feature map (e). \label{fig:attention}}

\Figure[t!][scale = 0.6]{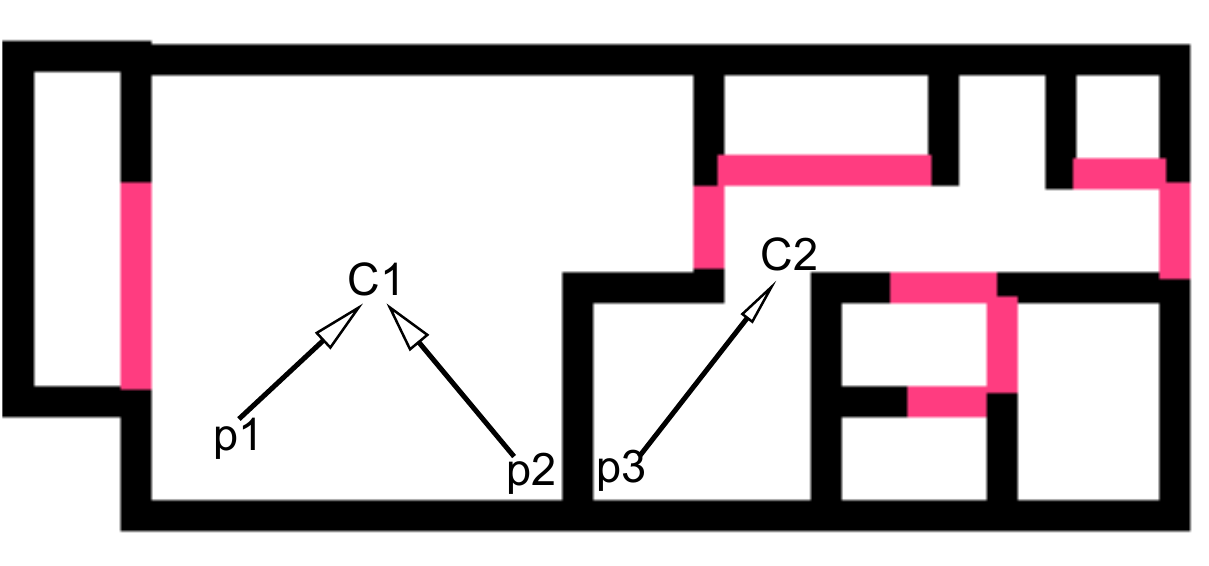}{Illustration of the offset prediction module in the Offset-Guided Attention. This module predicts the offset from each pixel to the center of room it belongs to. The correlation between the two pixels is measured by the distance of the them after moving along according to the offset towards the room center.
\label{fig:fig4}
}


\section{Method}

\subsection{Overview}
Fig 2 shows the overall architecture of our network. We adopt the ResNet-101~\cite{b19} backbone to extract feature maps from the input image. After the feature extraction, the prediction head of the network contains four branches: boundary prediction, room prediction, Offset-Guided Attention, and Feature-Fusion Attention. Boundary prediction (Fig~\ref{fig:overall} (4)) is to predict of boundary categories, i.e., walls and doors. Room prediction (Fig~\ref{fig:overall} (5)) is to predict the semantic category of the pixels within rooms. Offset-Guided Attention (Fig~\ref{fig:attention}) refers to the prediction of the vertical and horizontal offset between the specific pixel $p$ and the geometrical center $c$ of the room $R$ it belongs to. Feature-Fusion Attention (Fig~\ref{fig:fusion}) aims to improve the consistency between the room and boundary predictions. In the following, we will detail the key modules including the boundary and room prediction in Section~\ref{sec:2branch}, Offset-Guided Attention in Section~\ref{sec:offset}, and Feature-Fusion Attention in Section~\ref{sec:fusion}.


\subsection{Boundary and Room Prediction}~\label{sec:2branch}
%
We adopt ResNet-101 with 8$\times$ down-sampling rate as the backbone network for feature extraction. Inspired by ~\cite{b1}, we use two individual branches on the top of the backbone for room and boundary prediction respectively, as shown in Fig~\ref{fig:overall}(4, 5) to improve their performances.  

\Figure[t!][scale=0.8]{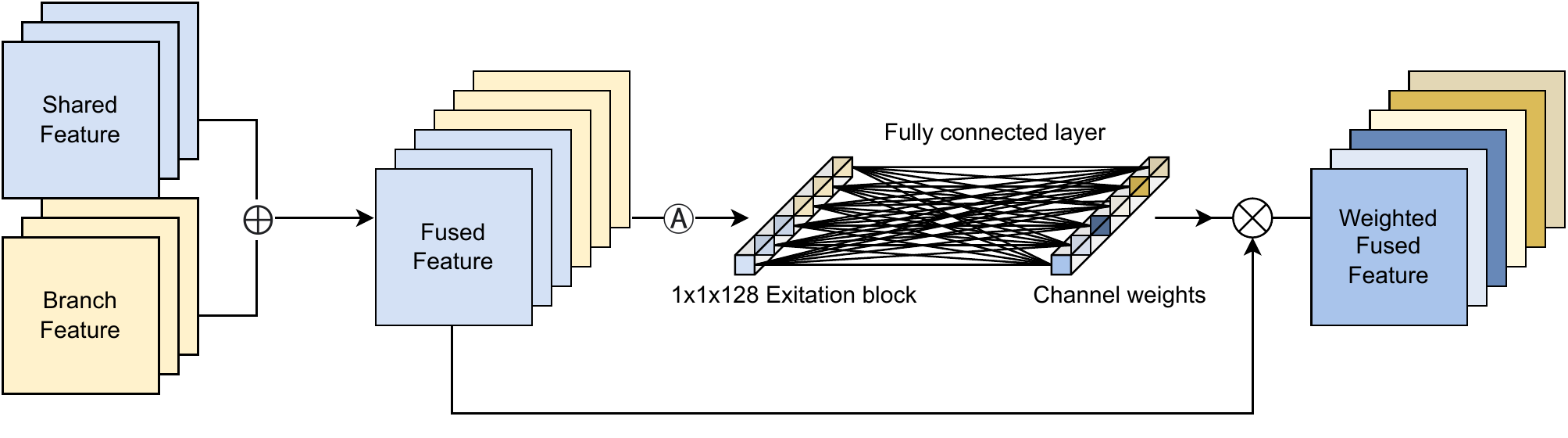}{Illustration of the Feature Fusion Attention module, where \textcircled{A} denotes average pooling layer, $\bigotimes$ denotes multiplication, and $\bigoplus$ denotes concatenation.\label{fig:fusion}} 
\subsection{Offset-Guided Attention Module}~\label{sec:offset} 
In the floor plan image, the semantic correlation between two pixels cannot be directly represented by their distance on the floor plan. As shown in Fig \ref{fig:fig4}, pixels that are close to each other (p2, p3) can be hardly related in terms of semantic category since they belong to different rooms, whereas those that are far away (p1, p2) are semantically correlated with each other since they belong to the same room. Inspired by the above observation, we propose an Offset-Guided Attention module to better measure the pixel correlation. 

Fig~\ref{fig:attention} shows the architecture of Offset-Guided Room Attention Module. First, as shown in figure~\ref{fig:attention} (a), the offset prediction module predicts the the offset $o_{p}$ from each pixel $p = (x, y)$ to the center of the room $c_p$ it belongs to, i.e., $o_{p}=c_p-p$. With equation
\begin{equation}
p'=p+o_p
\label{equ1}
\end{equation}
denoting the offset-guided shifted position of $p$, we can measure the correlation $A_{i,j}$ of two pixels $p_i, p_j$ according to the Equation~\ref{equ2}.
\begin{equation}
A_{i,j} = || p'_i - p'_j ||_2^2
\label{equ2}
\end{equation}
If the pixels $p_i, p_j$ are close to each other after the offset-guided shifting, then they are strongly correlated with each other, and hence have better chance to be in the same room and also share the same semantic category. Then, $A$ further serves as the attention map of our Offset-Guided Attention module. In Fig~\ref{fig:attention} (c), we illustrate the mutual attention map of the pixels in one row $I_i$ of the image $I$. This sub attention map has the shape $w\times w$ where $w$ indicates the width of the image. It is the symmetric matrix where the diagonal values indicate the correlation to the pixel itself. In addition, the attention of nearby pixels (in the same room) are significantly larger than others that are located in different room, and the sharp change of attention values in Fig~\ref{fig:attention} (c) manifest the effectiveness of our attention map $A$ to distinguish the pixels within the same room or are in different rooms. 

Then, the attention map $A$ is normalized by softmax to make sure that $\sum_j A_{i,j}$, i.e., the sum of attention values of all other pixels $j$ to the target pixel $i$, to be 1. Then the features from the room prediction branch $f_j$ are weighted summed according to $A$, and produce the enhanced feature $\hat{f_i}$ for the target pixel $i$, as shown in Equation~\ref{sum}.
\begin{equation}
\hat{f}_i=\sum_j A_{i,j}f_j
\label{sum}
\end{equation}
At last, the enhanced feature $\hat{f}_i$ is concatenated with $f_i$ for the final semantic prediction. Note that here we take the pixel $i$ as an example, and in our approach, all the pixels are operated in the same way.

To improve the memory and calculation efficiency, we adopt crisscross attention technique\cite{b12}, where only the correlation between the target pixel $i$ and other pixels $j$ on its horizontal and vertical axis contribute to the attention map. Following \cite{b12}, this module is applied for two times successively in order to simulate the full attention. Please refer to \cite{b12} for more details.

\subsection{Feature Fusion Attention Module}~\label{sec:fusion}
As discussed in Section~\ref{sec:introduction}, the predictions of room and boundary might be inconsistent with each other, since the two branches introduced in Section~\ref{sec:2branch} predict these two kinds of categories individually, as shown in Fig~\ref{fig:fig1}. To address this issue, we propose a Feature Fusion Attention module to fuse the features of the two branches with an additional shared branch, and enhance the prediction consistency of room and boundary branches. 
Fig~\ref{fig:fusion} shows the architecture of Feature Fusion Attention module. 
First, the feature map from the two branches $f_1, f_2 \in R^{d\times h\times w}$ are concatenated in the channel dimension as $f=f_1\oplus f_2 \in R^{2d\times h\times w}$, and a global average pooling layer reduces the size of $f$ to be $\in R^{2d\times 1\times 1}$. Then one fully connected layer learns a channel-wise attention map $A_c$ of size $2d\times 2d$, indicating how each channel correlates to others, and outputs a weight feature $w \in R^{2d\times 1\times 1}$, one element for each channel. At last, the enhanced feature map is derived with Equation~\ref{equ:fw}. 
\begin{equation}
\hat{f} = f \otimes w
\label{equ:fw}
\end{equation}
where the $\otimes$ denotes the element-wise multiplication. 

To summarize, the Feature Fusion Attention module leverage the shared features of the boundary and room prediction branches and feed the mutual information between them to improve their prediction consistency.

\subsection{Loss function}

For both the boundary and room prediction, we use weighted softmax cross entropy loss with specific weight for different categories, as shown in Equation~\ref{equ:softmax}. 
\begin{equation}
    \mathcal{L}_s=\sum_{i=1}^{C}-w_i\mathbbm{1}(y_i)\log(p_i)
\label{equ:softmax}
\end{equation}
where $C$ is the number of categories, $w_i$ is the weight applied to the specific category, $\mathbbm{1}(y_i)$ is an indicator function and equals to 1 when the ground truth category of $y$ is the category $i$. $p_i$ is the predicted possibility for the category normalized by the Softmax function. $w_i$ is derived by calculating the percent of pixels of the specific category in the training set:
\begin{equation}
    w_i = \frac{V_i}{V_{total}}
\end{equation}
where $V_i$ is number of pixels of the specific category, and $V_{total}$ is the total number of pixels. 

For the offset prediction, we use $L_1$ loss to measure the distance of two pixels after offset-guided shifting, as shown in Equation~\ref{equ:offset}. 
\begin{equation}
    \mathcal{L}_o=|g_x - o_x| + |g_y - o_y|
    \label{equ:offset}
\end{equation}
where $g_x$ and $g_y$ are the offset ground truth in x and y axis, and $o_x$ and $o_y$ are the predicted offset in x and y axis. 

The overall loss to train the network is $\mathcal{L}_s+\mathcal{L}_o$. 

\begin{figure*}
\begin{center}
\subfigure[Input]{\includegraphics[width=2.4cm]{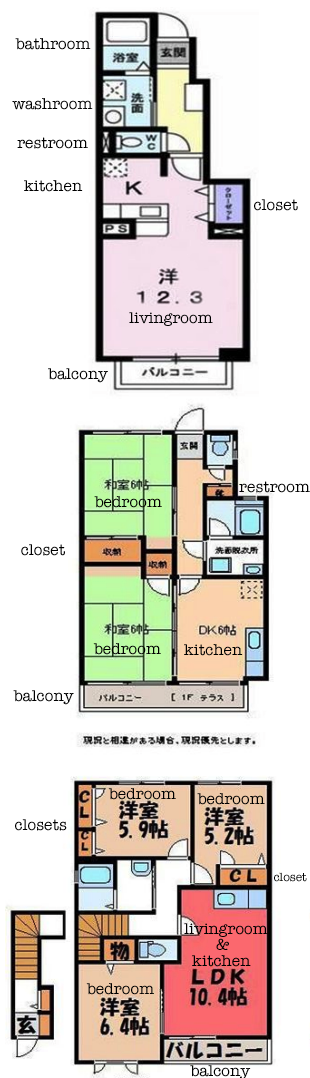}}
\subfigure[Ground Truth]{\includegraphics[width=2.3cm]{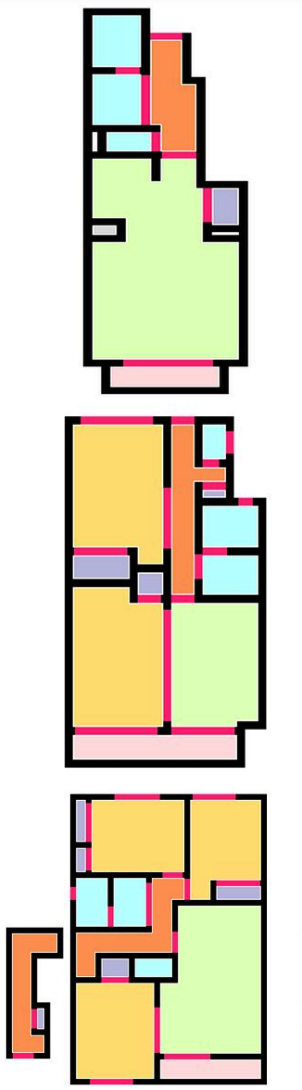}}
\subfigure[Ours]{\includegraphics[width=2.5cm]{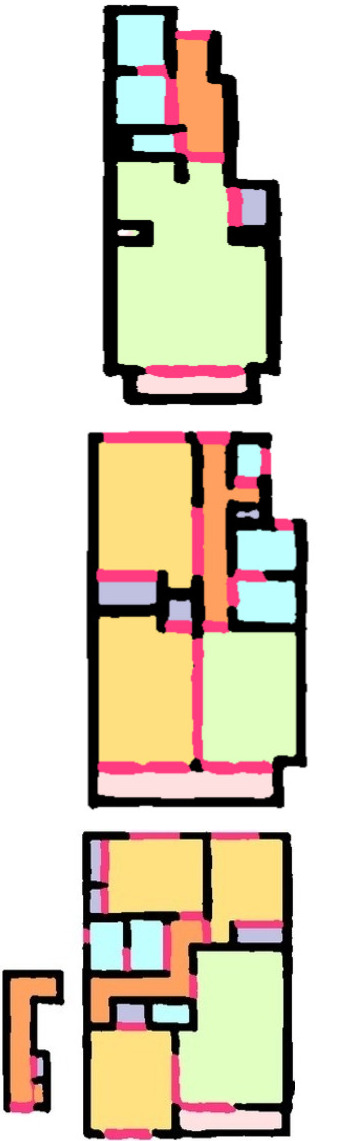}}
\subfigure[DeepFloorPlan]{\includegraphics[width=2.32cm]{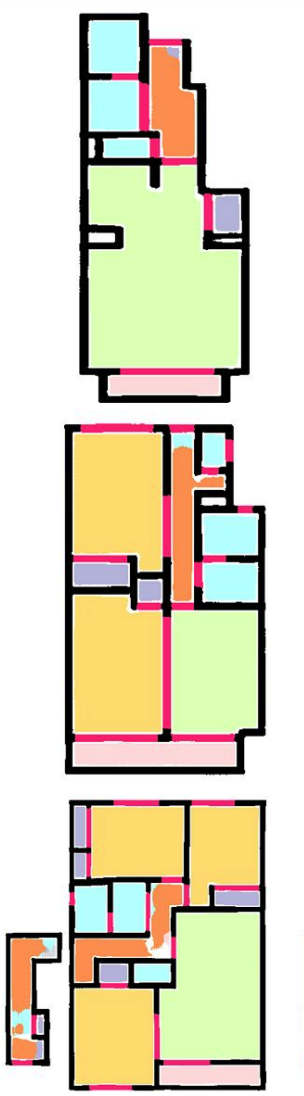}}
\subfigure[R2V]{\includegraphics[width=2.31cm]{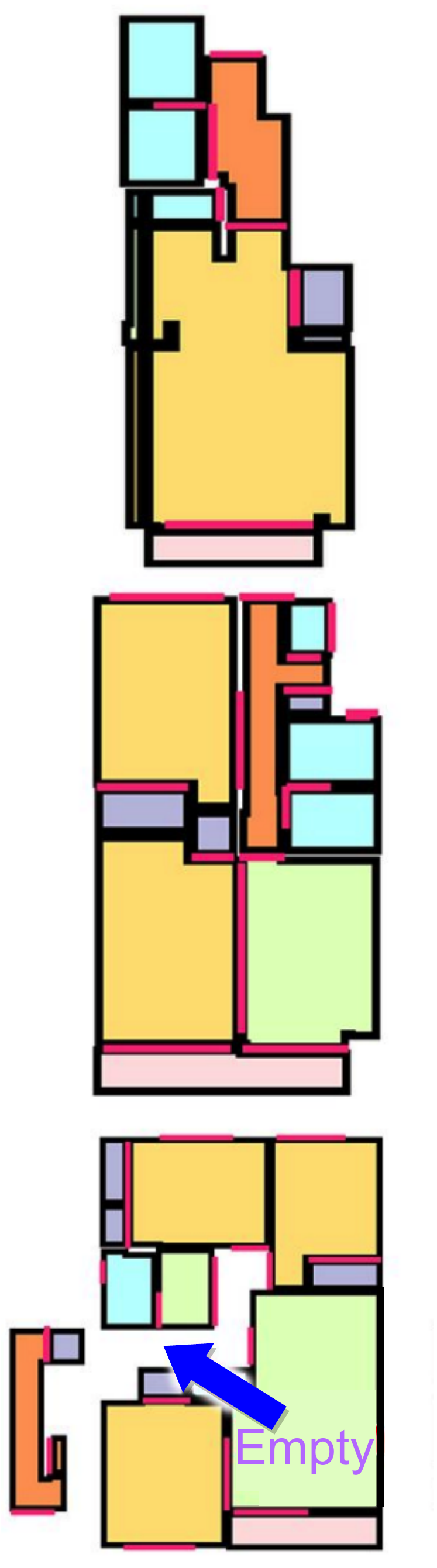}}
\subfigure[DeepLabV3+]{\includegraphics[width=2.32cm]{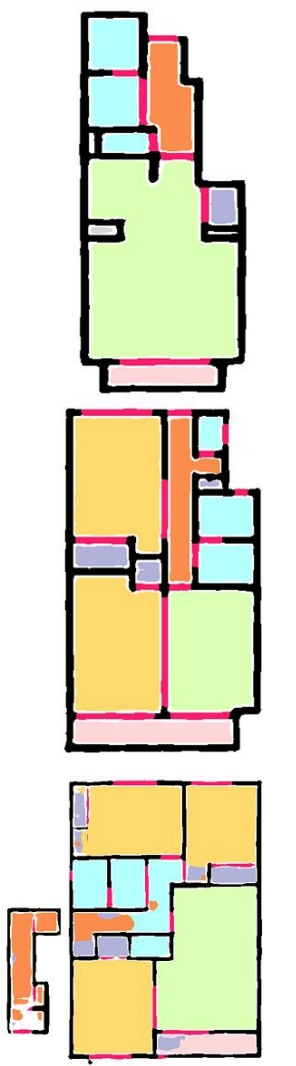}}
\subfigure[PSPNet]{\includegraphics[width=2.175cm]{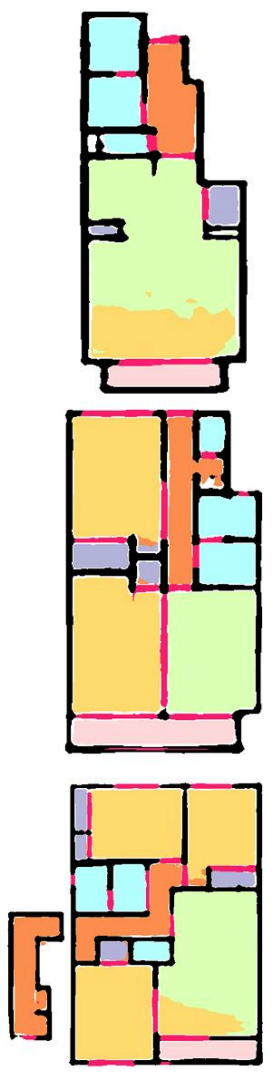}}
\caption{Visual comparison of our result(c) with existing works(d-g), including Deep Floor Plan\cite{b1}, R2V(Raster-to-Vector)\cite{b4}, DeepLabV3+\cite{b20}, and PSPNet\cite{b21}, on the R2V dataset. Note that the English labels in (a) do not exist in the original image, and are added for reader's reference.}
\label{fig:compare}
\end{center}
\end{figure*}

\section{Experiments}
\subsection{Datasets}
We evaluate our Offset-Guided Attention Networks on two datasets that were commonly used for floor plan recognition, the R3D and R2V dataset~\cite{b13, b4}. The R3D dataset consists of 214 images from \cite{b13}, and the R2V dataset consists of 815 images from \cite{b4}. To have a fair comparison with existing works, we follow the same train-test split as~\cite{b1}. 

\subsection{Implementation Details}
Our network is trained on a single NVIDIA Tesla P100-PCIE GPU and is trained for 7k iterations in total. 
We adopted a fixed initial learning rate of $0.01$ and used a weight decay of rate $0.0002$. Besides, we use SGD with Nesterov Momentum and update parameters. 
The training is conducted in two steps. First, we initialize our network using the ResNet model pretrained on ImageNet and optimize the offset prediction module with other weights frozen. Then, we optimize other modules in our network with offset prediction branch frozen. In inference, we derive the final result by a single forward pass, without any post processing like existing works~\cite{b1, b3}.

\subsection{Metrics}

Following the existing works~\cite{b1, b4}, we adopt $overall accuracy$ and $class accuracy$ as the metrics to evaluate the performance quantitatively as shown in Equation~\ref{equ:acc} and ~\ref{equ:clsacc}.
\begin{equation}
    overall\_acc = \sum^{i}_{j=0} \frac{C(j)}{N(j)}, 
\label{equ:acc}
\end{equation}
\begin{equation}
    class\_acc(j) = \frac{C(j)}{N(j)}
\label{equ:clsacc}
\end{equation}
where $overall\_acc$ denotes overall accuracy, $class\_acc$ denotes class accuracy of a specific class $j$, $i$ is the number of categories, $C$ is the correctly predicted number of pixels of specific category $j$, and $N$ is the total number of pixels of specific category $j$. 

In addition, we also adopt $m\_IoU$(mean Intersection over Union) as an additional metric, defined as the following equation:
\begin{equation}
    m\_IoU = \frac{\sum^{n}_{j=0} \frac{C_j}{P_j + G_j - C_j}}{n}, 
\label{equ:miou}
\end{equation}
where $n$ denotes total number of categories, $C_j$ is the correctly predicted number of pixels in certain category, $P_j$ is the total number of pixels predicted in certain category, and $G_j$ is total number of pixels in the ground truth of certain category.


\newcommand{\wideunderline}[2][2em]{%
  \underline{\makebox[\ifdim\width>#1\width\else#1\fi]{#2}}%
}
\subsection{Comparison with the Existing Works}

We compare our network with existing floor plan recognition approaches Raster-to-Vector \cite{b4} and Deep Floor Plan \cite{b1}, as well as general semantic segmentation approaches PSPNet~\cite{b21} and DeepLabV3+~\cite{b20}. 

Fig \ref{fig:compare} (c, e) show visual comparisons of our result and Raster-to-Vector. Comparing with the ground truth (b), Raster-to-Vector (e) fails to determine the room type of a major room in the top row. 
In addition, their model fails to recognize some room regions and leaves it empty. In contrast, our model shows perfect room type prediction and accurate recognition of boundary categories. As shown in Table \ref{tab:compare} ``Raster-to-Vector'', our result has superior performances in all categories and has significantly higher overall accuracy.

\begin{table*}
    \begin{center}
    \begin{tabular}{|c||c||c|c|c|c|c||c|c|c|c|}
        \hline
        \multirow{2}{3.4em}{}& \multirow{2}{6em}{}& \multicolumn{5}{|c||}{{R2V}} & \multicolumn{4}{|c|}{{R3D}}\\
        \cline{3-11}
        & & Ours & Raster-to-Vector & DeepLabV3+ & PSPNet &  DFP & Ours & DeepLabV3+ & PSPNet & DFP \\
        \cline{2-11}
        & \emph{\textbf{overall accuracy}} & \textbf{0.93} & 0.84 & 0.88 & 0.88 & 0.89 & \textbf{0.91} & 0.85 & 0.84 & 0.89 \\
        \cline{2-11}
        & \emph{\textbf{mean IoU}} & \textbf{0.77} & N/A & 0.69 & 0.70 & 0.74 & \textbf{0.74} & 0.50 & 0.50 & 0.63 \\
        \hline
        \multirow{8}{3.4em}{\emph{class accuracy}} & wall & \textbf{0.90} & 0.53 & 0.80 & 0.84 & 0.89 & 0.95 & 0.93 & 0.91 & \textbf{0.98}\\
        \cline{2-11}
        & door \& window & \textbf{0.89} & 0.58 & 0.72 & 0.76 & \textbf{0.89} & \textbf{0.83} & 0.60 & 0.54 & \textbf{0.83}\\
        \cline{2-11}
        & closet & \textbf{0.88} & 0.78 & 0.78 & 0.80 & 0.81 & \textbf{0.71} & 0.24 & 0.45 & 0.61\\
        \cline{2-11}
        & bathroom & \textbf{0.92} & 0.83 & 0.90 & 0.90 & 0.87 & \textbf{0.90} & 0.76 & 0.70 & 0.81\\
        \cline{2-11}
        & living room & \textbf{0.94} & 0.72 & 0.85 & 0.83 & 0.88 & \textbf{0.93} & 0.76 & 0.76 & 0.87\\
        \cline{2-11}
        & bedroom & \textbf{0.96} & 0.89 & 0.82 & 0.86 & 0.83 & \textbf{0.89} & 0.56 & 0.55 & 0.75\\
        \cline{2-11}
        & hall & \textbf{0.84} & 0.64 & 0.55 & 0.78 & 0.68 & \textbf{0.87} & 0.72 & 0.61 & 0.59\\
        \cline{2-11}
        & balcony & \textbf{0.91} & 0.71 & 0.87 & 0.87 & 0.90 & \textbf{0.83} & 0.08 & 0.41 & 0.44\\
        \hline
    \end{tabular}
    \end{center}
    \caption{Quantitative comparison with Raster-to-Vector\cite{b4}, DeepLabv3+\cite{b20}, PSPNet\cite{b21}, and Deep Floor Plan(DFP)\cite{b1} on both R2V and R3D. Our method produces superior results in both datasets and has higher accuracy in almost all sub-classes. We did not report the m\_IoU of Raster-to-Vector model since m\_IoU metric is not conducted in their original paper\cite{b4} and the model's relatively weak performance under accuracy metric implies that is has relatively low m\_IoU.}
    \label{tab:compare}
\end{table*}

\begin{figure*}
\begin{center}
\subfigure[Input]{\includegraphics[width=2.74cm]{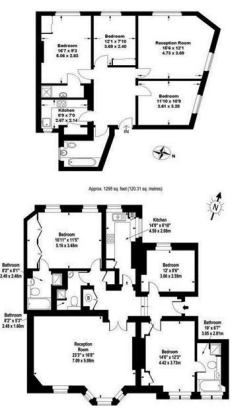}}
\subfigure[Ground Truth]{\includegraphics[width=2.47cm]{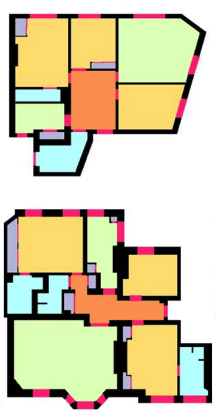}}
\subfigure[Ours]{\includegraphics[width=2.461cm]{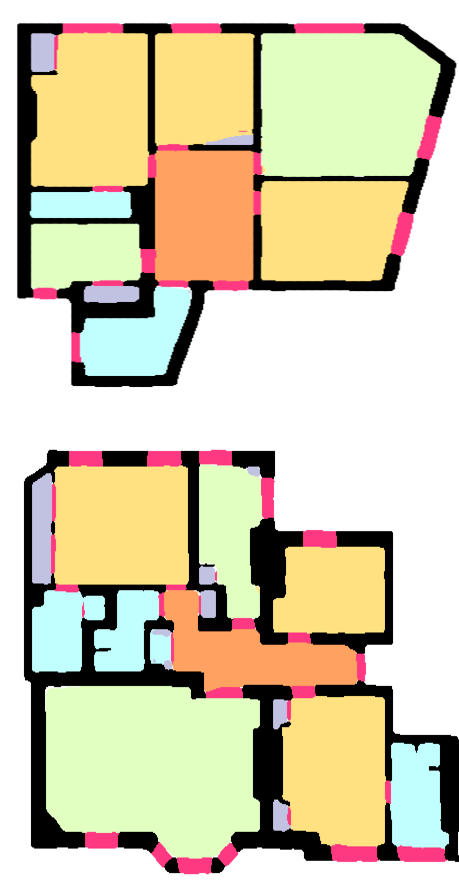}}
\subfigure[DeepFloorPlan]{\includegraphics[width=2.5cm]{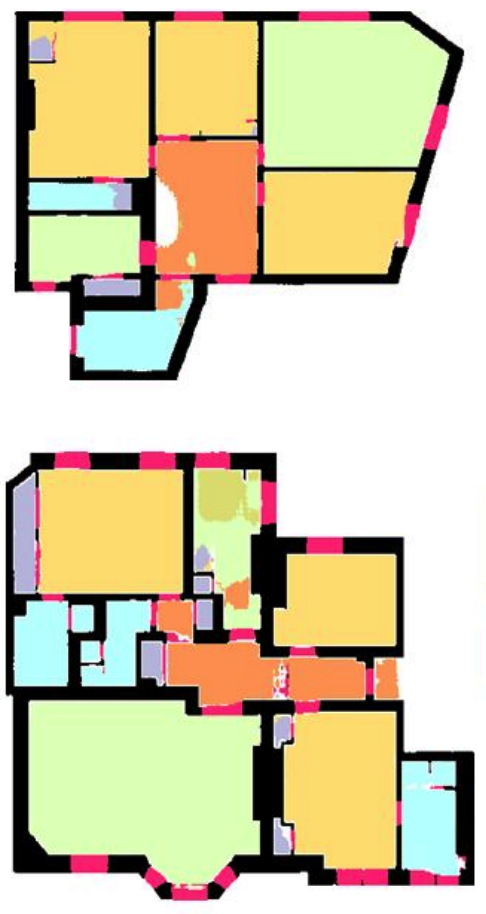}}
\subfigure[DeepLabV3+]{\includegraphics[width=2.46cm, trim={0cm 0cm 0.03cm 0.03cm},clip]{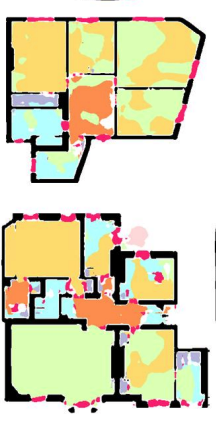}}
\subfigure[PSPNet]{\includegraphics[width=2.48cm]{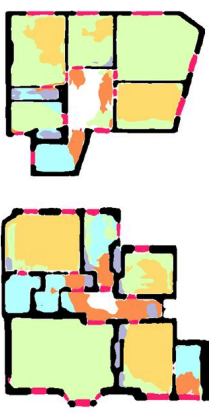}}
\caption{Visual comparison of our result(c) with existing works(d-g), including Deep Floor Plan\cite{b1}, R2V(Raster-to-Vector)\cite{b4}, DeepLabV3+\cite{b20}, and PSPNet\cite{b21}, on the R3D dataset}
\label{fig:r3dcompare}
\end{center}
\end{figure*}

Next, we compare our result with general segmentation networks including DeepLabV3+\cite{b20} and PSPNet\cite{b21}. We utilize the author-released code to produce their results. First, as shown in Fig \ref{fig:compare} (c, f, g), in the top and bottom case, both DeepLabV3+ and PSPNet are not able to predict one or more rooms to be the unified category. Besides, there are inconsistencies between their boundary and room predictions, such as some nearby room categories are dilated into each other. In contrast, our network produces the uniform and consistent results within rooms. 
Also, Fig \ref{fig:r3dcompare} shows our network's ability to produce superior results in the R3D dataset over the existing works. 
Table \ref{tab:compare} ``DeepLabV3+'' and ``PSPNet'' show the quantitative comparison between our results and results of the two existing works on the R2V and R3D datasets. As shown in the table, our network produces significantly superior results in overall accuracy and all sub-classes in comparison with the other two networks in both R2V and R3D datasets. This result explicitly shows the superiority of our network over the general semantic segmentation networks, and also demonstrates the significant difference between the floor plan recognition task and the general semantic segmentation task. 

\begin{figure*}
    \centering
    \begin{subfigure}[Input]{
    \includegraphics[width=1.63cm, trim={4cm 4.5cm 3.5cm 0cm},clip]{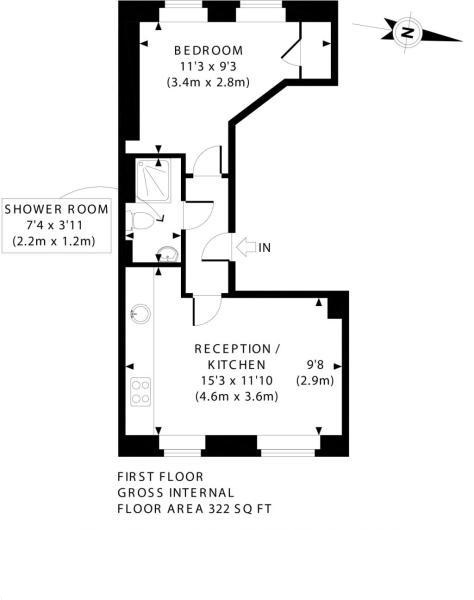}
    \includegraphics[width=2.91cm, trim={0cm 3cm 0 0}, clip]{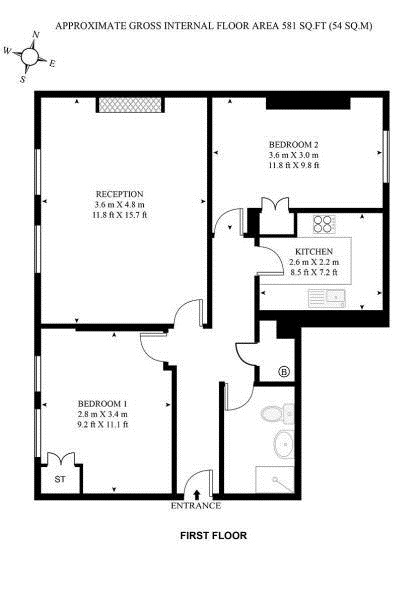}
    \includegraphics[width=2.85cm, trim={0cm 5cm 0cm 0cm},clip]{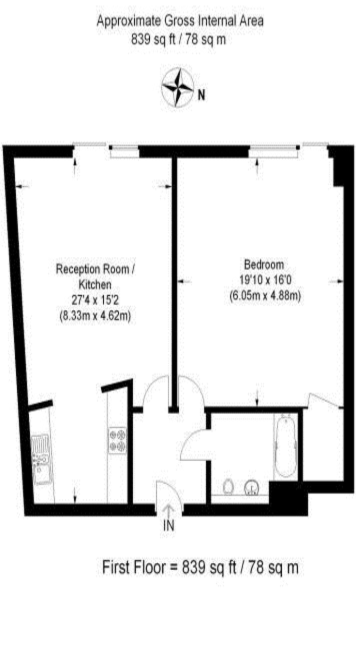}
    \includegraphics[width=2.35cm, trim={1cm 2.5cm 1.5cm 3cm},clip]{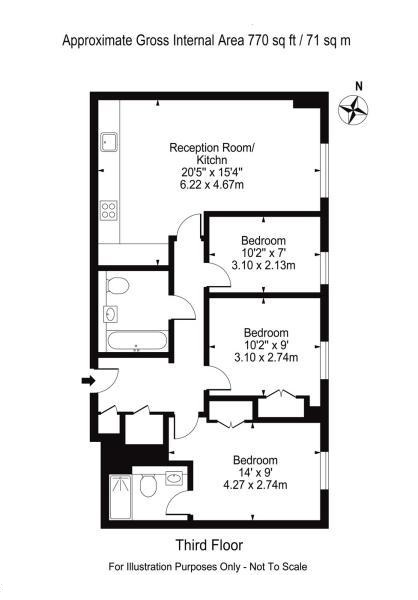}
    \includegraphics[width=3.1cm, trim={0.5cm 0cm 1cm 0cm},clip, angle =90 ]{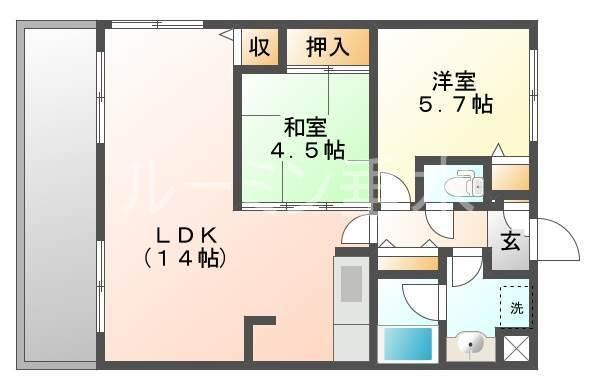}
    \includegraphics[width=3cm, trim={0.2cm 4.5cm 0cm 5.5cm}, clip, angle =90 ]{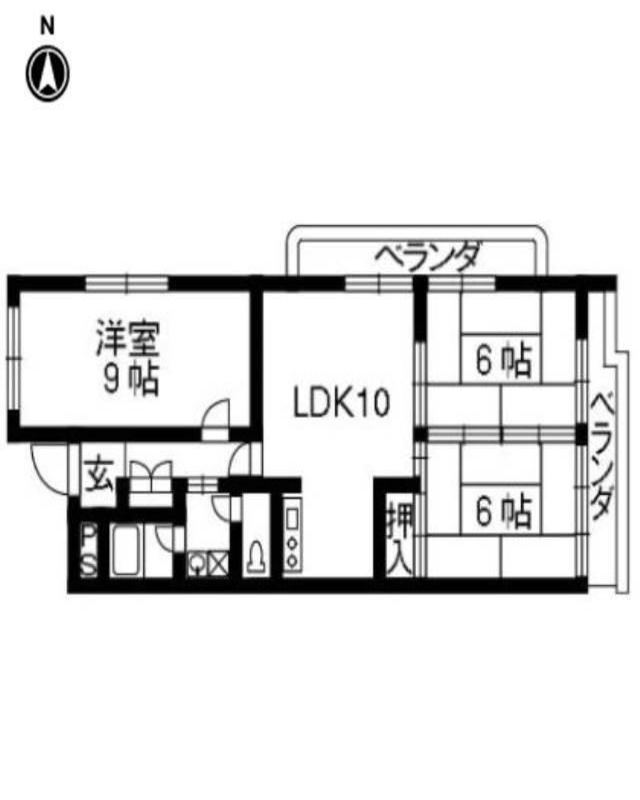}
    \includegraphics[width=3.2cm, trim={1cm 4cm 0cm 6.2cm},clip, angle =90 ]{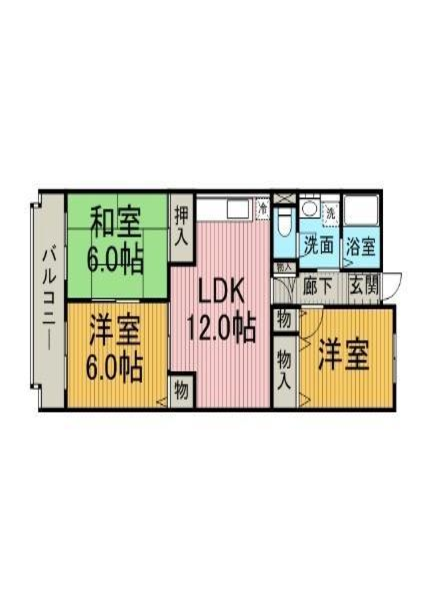}
    }
    \end{subfigure}
    \begin{subfigure}[Ours]{
    \includegraphics[width=1.63cm, trim={4cm 4.5cm 3.5cm 0cm},clip]{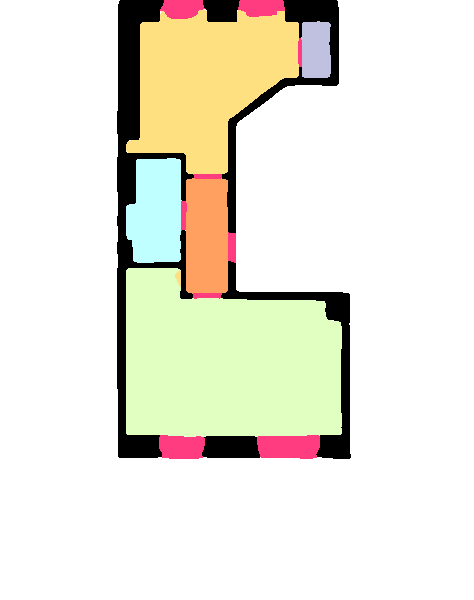}
    \includegraphics[width=2.91cm, trim={0cm 3cm 0 3cm}, clip]{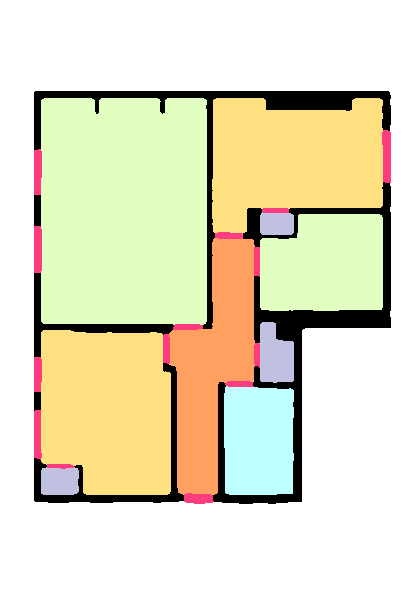}
    \includegraphics[width=2.85cm, trim={0cm 5cm 0cm 3,5cmcm},clip]{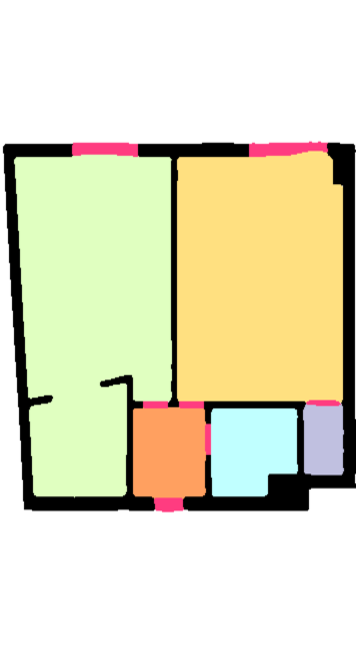}
    \includegraphics[width=2.35cm, trim={1cm 2.5cm 1.5cm 3cm},clip]{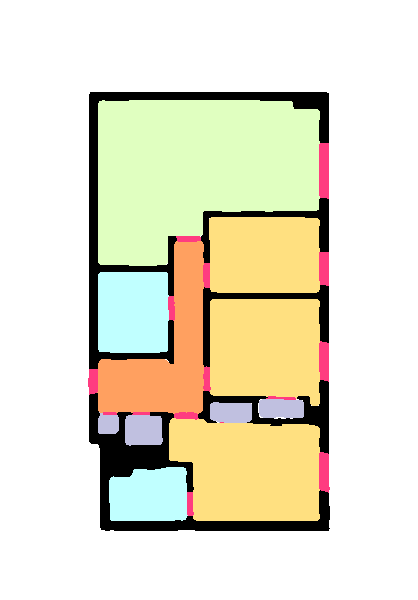}
    \includegraphics[width=3.1cm, trim={0.5cm 0cm 1cm 0cm},clip, angle =90 ]{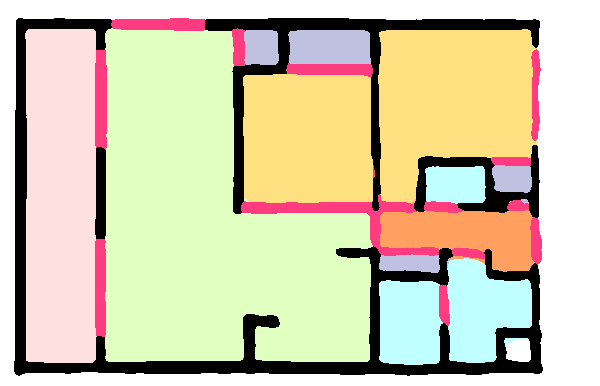}
    \includegraphics[width=3cm, trim={0.2cm 4.5cm 0cm 5.5cm}, clip, angle =90 ]{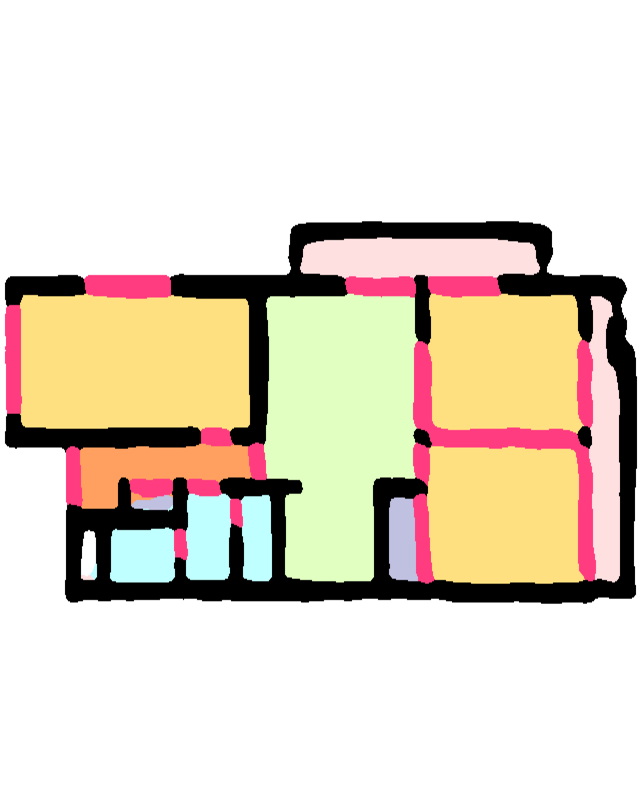}
    \includegraphics[width=3.2cm, trim={1cm 4cm 0cm 6.2cm},clip, angle =90 ]{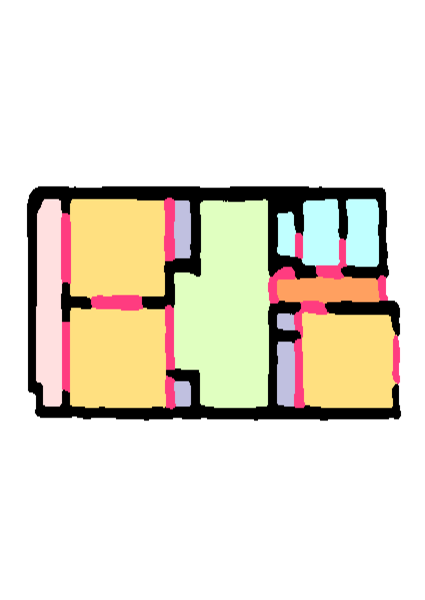}
    }
    \end{subfigure}
    \begin{subfigure}[Ground Truth]{
    \includegraphics[width=1.63cm, trim={4cm 4.5cm 3.5cm 0cm},clip]{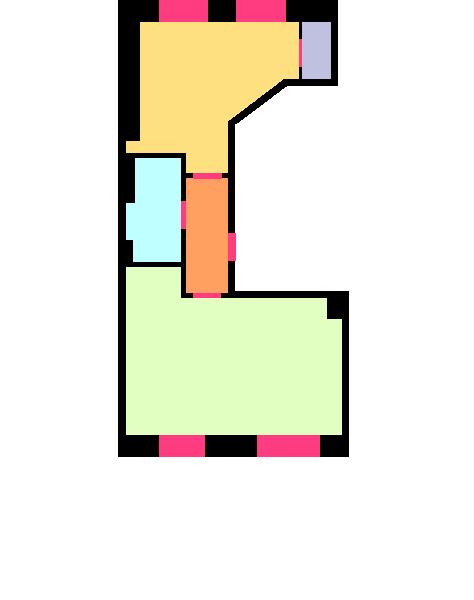}
    \includegraphics[width=2.91cm, trim={0cm 3cm 0 3cm}, clip]{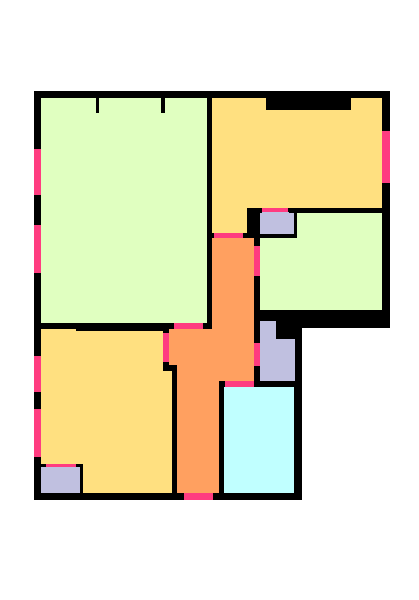}
    \includegraphics[width=2.85cm, trim={0cm 5cm 0cm 3.5cm},clip]{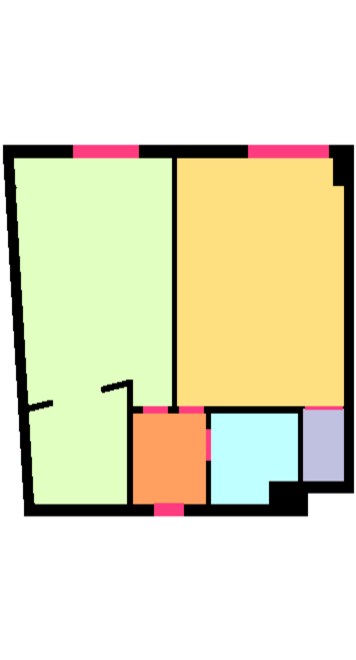}
    \includegraphics[width=2.4cm, trim={1cm 2.5cm 1.5cm 3cm},clip]{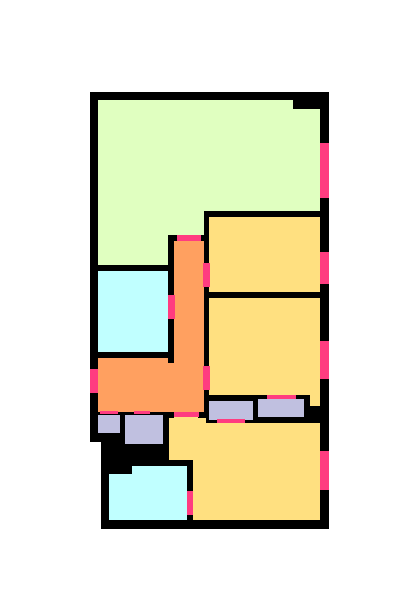}
    \includegraphics[width=3.1cm, trim={0.5cm 0cm 1.5cm 0cm},clip, angle =90 ]{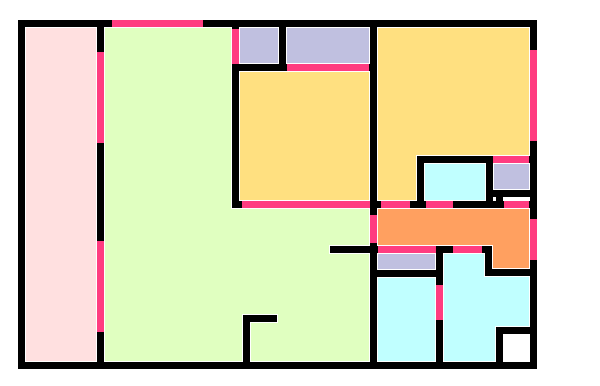}
    \includegraphics[width=3cm, trim={0.2cm 4.5cm 0cm 5.5cm}, clip, angle =90 ]{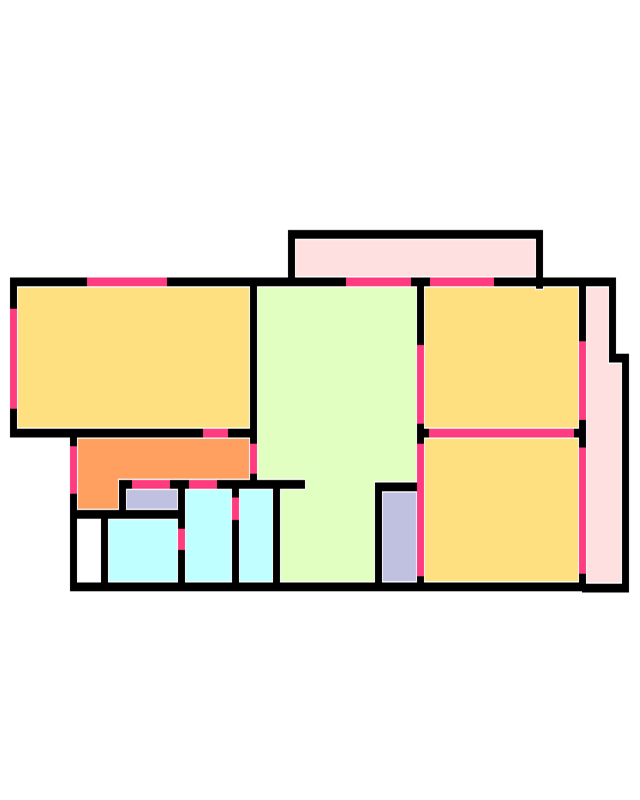}
    \includegraphics[width=3.2cm, trim={1cm 4cm 1cm 6.2cm},clip, angle =90 ]{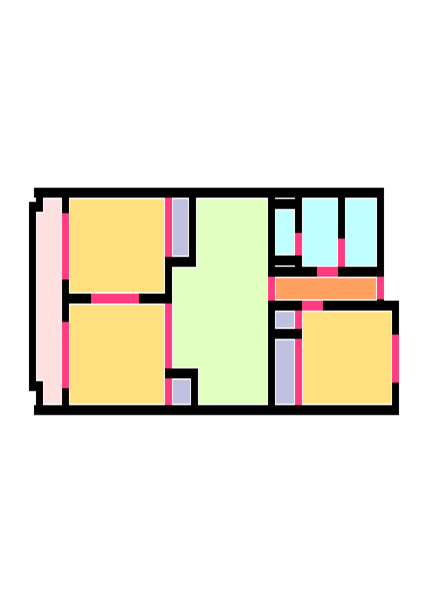}
    }
    \end{subfigure}
    \caption{More results produced by our network on the R3D and R2V dataset}
    \label{fig:moreresults}
\end{figure*}

We further compare our work with a recent floor plan recognition approach Deep Floor Plan~\cite{b1}. Still, we utilize the author-released model and adopt the same experimental setting as in their paper. 
As shown in Fig \ref{fig:compare} (d), 
in the middle case, there is a blue part predicted inside the orange room, indicating that their approach is incapable to have the consistent room-type prediction, and can make the wrong prediction when the room has the irregular shape. In addition, some predictions are dilated to the nearby rooms as shown in the bottom row, leading to inconsistent predictions with a single room. On the contrary, as shown in Fig \ref{fig:compare} (c), our model is able to generate consistent and correct predictions with our Offset-Guided Attention Mechanism combined with Feature Fusion Attention module, no matter what the shape or size of the room is. In R3D dataset shown as Fig  \ref{fig:r3dcompare} (c, d), our results achieve the superior room uniformity and consistency over Deep Floor Plan~\cite{b1}. Quantitatively, according to Table \ref{tab:compare}, our model has higher accuracy for almost all sub-classes for both R2V and R3D datasets. 


\subsection{More results}
We further proceed to analyse our network's ability in various specific cases. We divide R3D floor plans into two types, irregular (first and third case) and rectangular (second and last case), and selected two cases of each type. As shown in Fig \ref{fig:moreresults} (a, b), our model is able to produce highly accurate and unified room predictions in both situations, manifesting the ability of our network in handling rooms with different shapes. Besides, our model is able to recognize walls of different thicknesses accurately. As for the R2V dataset, it is mostly composed of rectangular floor plans, so we select the challenging cases that have complicated layouts and contain multiple rooms that look similar. Fig \ref{fig:moreresults} (c, d) shows that our model can consistently generate precise and concrete predictions on all of them.

\section{Ablation study}
\subsection{Offset-Guided Attention Module}
\begin{figure}
    \centering
    \subfigure[Input image]{\includegraphics[width=4cm, trim={0 3cm 0 3cm},clip]{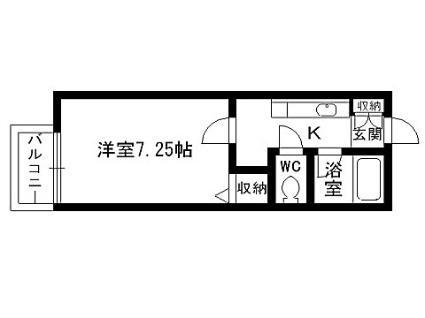}}\
    \subfigure[Ground truth]{\includegraphics[width=4cm, trim={0 3cm 0 3cm},clip]{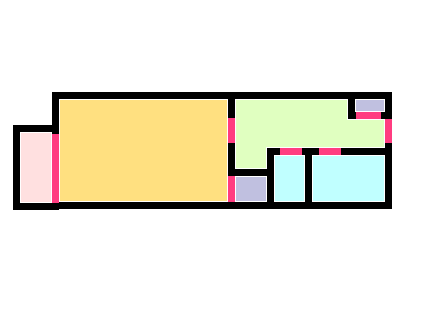}}
    \subfigure[Ours without FFA]{\includegraphics[width=4cm, trim={0.5 3cm 0 3cm},clip]{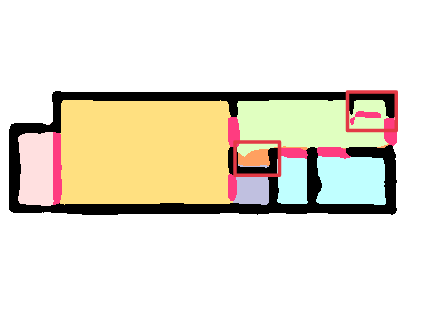}}
    \subfigure[Ours]{\includegraphics[width=4cm, trim={0.3cm 3cm 0 3cm},clip]{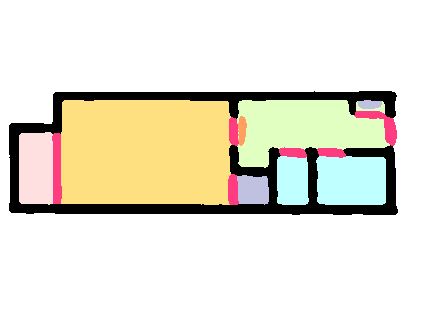}}
    \caption{Results produced by our model with and without Feature Fusion Attention(FFA) module. The red boxes denote the disagreement of the two branches without the Feature Fusion Attention module.}
    \label{fig:ablation}
\end{figure}

\begin{table}[]
    \centering
    \begin{tabular}{c||c|c|c}
        \hline
        & Ours & Without OGA & Without FFA\\
        \hline
        \textbf{overall acc} & 0.93 & 0.91\textbf{(-0.02)} &  0.92\textbf{(-0.01)}\\
        \hline
        wall & 0.90 & 0.93\textbf{(+0.03)} & 0.90 \\
        \hline
        door \& window & 0.89 & 0.86\textbf{(-0.03)} & 0.84\textbf{(-0.05)} \\
        \hline
        closet & 0.88 & 0.87\textbf{(-0.01)} & 0.87\textbf{(-0.01)} \\
        \hline
        bathroom & 0.92 & 0.91\textbf{(-0.01)} & 0.91\textbf{(-0.01)} \\
        \hline
        living room & 0.94 & 0.90\textbf{(-0.04)} & 0.90\textbf{(-0.04)} \\
        \hline
        bedroom & 0.96 & 0.95\textbf{(-0.01)} & 0.96 \\
        \hline
        hall & 0.84 & 0.82\textbf{(-0.02)} & 0.87\textbf{(+0.03)}\\
        \hline
        balcony & 0.91 & 0.86\textbf{(-0.05)} & 0.91\\
        \hline
    \end{tabular}
    \caption{Comparison of results produced by our network with both modules, without Offset-Guided Attention(OGA) module, and without Feature Fusion Attention(FFA) module on the R2V dataset. }
    \label{tab:ablation}
\end{table}
First, we evaluate the effectiveness of Offset-Guided Attention module on floor plan recognition task. 
As shown in Table \ref{tab:ablation} and Fig~\ref{fig:ablation}, all of the room categories attain inferior accuracy when the Offset-Guided Attention module is disabled, whereas the results of the boundary categories are not largely impacted. 
Notably, the performances of balcony, living room, and hall prediction decrease more significantly than the other categories since they are typically in either irregular shapes or have large areas. Our Offset-Guided Attention module helps to handle the irregular-shaped rooms by fusing the multiple predictions of the same room, and improve the large-area room performance by increasing the network's receptive field. 

\subsection{Analysis of Feature Fusion Attention module}
Then, we evaluate the effectiveness of our Feature Fusion Attention module. Similarly, we train and test our network without Feature Fusion Attention module on the R2V dataset and analyze the results. As shown in Fig \ref{fig:ablation} (c), without Feature Fusion Attention module, some orange regions dilate into the green room. In addition, the room on the top right corner is correctly predicted by the boundary prediction but not recognized by room prediction branch. Such kind of disagreement between the two branches harms the overall accuracy and also the visual quality. The above issues are addressed by our Fusion Attention module as shown in Fig \ref{fig:ablation} (d). Next, we study the quantitative results produced by the network with and without Feature Fusion Attention 
Table \ref{tab:ablation} shows quantitative comparisons. From the table, the Feature Fusion Attention module improves most boundary and door sub-class accuracy and improved overall accuracy by 1\%, manifesting the effectiveness of this module.

\subsection{More ablation studies}
At last, we conduct additional evaluations on some designs of our proposed modules. To begin with, in the Offset-Guided Attention module, we study the effectiveness of the softmax normalization and the room-aware attention mechanism. Without the softmax normalization, the sum of attention values of all other pixels in the image to the target pixel is not necessarily to be 1, making the feature aggregation unstable. As shown in Table \ref{tab:ablation2}, the overall accuracy drops by 5\% when softmax normalization is absent.    
To validate the effectiveness of the room-aware attention mechanism, we use an alternative attention mechanism based on the absolute distance between two pixels. As shown in Table \ref{tab:ablation2} ``Absolute Distance'', the overall accuracy has a 2\% decrease when Offset-Guided Attention is replaced by absolute distance based attention, illustrating the superiority of our Offset-Guided room attention module. 

Then, we evaluate several key components in our Feature Fusion Attention module, including the effectiveness of the Feature Fusion Attention module in each branch and average pooling. Table \ref{tab:ablation3} shows results when a specific part is absent or altered. Notably, the absence of the Feature Fusion Attention module in either room or boundary branch results in the general performance drop in multiple sub-classes, implying that the FFA module only expresses its functionality when applied on both branches, since this module is designed to enhance the prediction consistency of the two branches. Max pooling layer yields comparable results as our network with average pooling, since both of the pooling strategies are able to extract a semantic-meaningful global feature from the feature map for the subsequent attention operations. 


\section{Conclusion}
In this work, we present a novel method for floor plan recognition task that has the following contributions: first, we design an Offset-Guided Attention module to encourage the consistency of predictions within a single room; second, we design a Feature Fusion Attention module that further enhances the performance by enhancing the prediction consistency between room and boundary branches. 
Further, we extensively evaluate our approach on two commonly used datasets R2V and R3D. Qualitative and quantitative evaluations show the superiority of our network over the existing works on floor plan recognition. In the future, we will extend the work to more tasks and practical applications such as floor plan instance segmentation and 3D reconstruction, et.al. 

\begin{table}[]
    \centering
    \begin{tabular}{c||c|c|c}
        \hline
        & Ours & Without softmax & Absolute Distance\\
        \hline
        overall acc & 0.93 & 0.88\textbf{(-0.05)} & 0.91\textbf{(-0.02)}\\
        \hline
        wall & 0.90 & 0.81\textbf{(-0.09)} & 0.93\textbf{(+0.03)}\\
        \hline
        door \& window & 0.89 & 0.81\textbf{(-0.08)} & 0.90\textbf{(+0.01)}\\
        \hline
        closet & 0.88 & 0.73\textbf{(-0.15)} & 0.83\textbf{(-0.05)}\\
        \hline
        bathroom & 0.92 & 0.84\textbf{(-0.08)} & 0.87\textbf{(-0.05)}\\
        \hline
        living room & 0.94 & 0.95\textbf{(+0.01)} & 0.91\textbf{(-0.03)}\\
        \hline
        bedroom & 0.96 & 0.92\textbf{(-0.04)} & 0.95\textbf{(-0.01)}\\
        \hline
        hall & 0.84 & 0.47\textbf{(-0.37)} & 0.75\textbf{(-0.09)}\\
        \hline
        balcony & 0.91 & 0.86\textbf{(-0.05)} & 0.89\textbf{(-0.02)}\\
        \hline
    \end{tabular}
    \caption{Additional ablation studies on Offset-Guided Attention module. ``Absolute Distance'' means applying a absolute-distance-based attention instead of Offset-Guided Attention in the module.}
    \label{tab:ablation2}
\end{table}

\begin{table}[]
    \centering
    \begin{tabular}{c||c|c|c|c}
        \hline
        & Ours & No R-FFA & No B-FFA & Max pooling\\
        \hline
        overall acc & 0.93 & 0.92\textbf{(-0.01)} & 0.92\textbf{(-0.01)} & 0.93\\
        \hline
        wall & 0.90 & 0.90 & 0.90 & 0.90\\
        \hline
        door \& window & 0.89 & 0.86\textbf{(-0.03)} & 0.87\textbf{(-0.02)} & 0.86\textbf{(-0.03)}\\
        \hline
        closet & 0.88 & 0.88 & 0.87\textbf{(-0.01)} &0.88\\
        \hline
        bathroom & 0.92 & 0.91\textbf{(-0.01)} & 0.90\textbf{(-0.02)} & 0.93\textbf{(+0.01)}\\
        \hline
        living room & 0.94 & 0.93\textbf{(-0.01)} & 0.92\textbf{(-0.02)} & 0.92\textbf{(-0.02)}\\
        \hline
        bedroom & 0.96 & 0.96 & 0.95\textbf{(-0.01)} & 0.96\\
        \hline
        hall & 0.84 & 0.84 & 0.83\textbf{(-0.01)} & 0.85\textbf{(+0.01)}\\
        \hline
        balcony & 0.91 & 0.89\textbf{(-0.02)} & 0.91 & 0.91\\
        \hline
    \end{tabular}
    \caption{Additional ablation studies on Feature Fusion Attention module. R-FFA represents the Feature Fusion Attention module applied to the room branch, and B-FFA represents the Feature Fusion Attention module applied to the boundary branch. Max pooling means replacing our average pooling layer with a max pooling layer.}
    \label{tab:ablation3}
\end{table}

\begin{IEEEbiography}[{\includegraphics[width=1in,height=1.25in,clip,keepaspectratio]{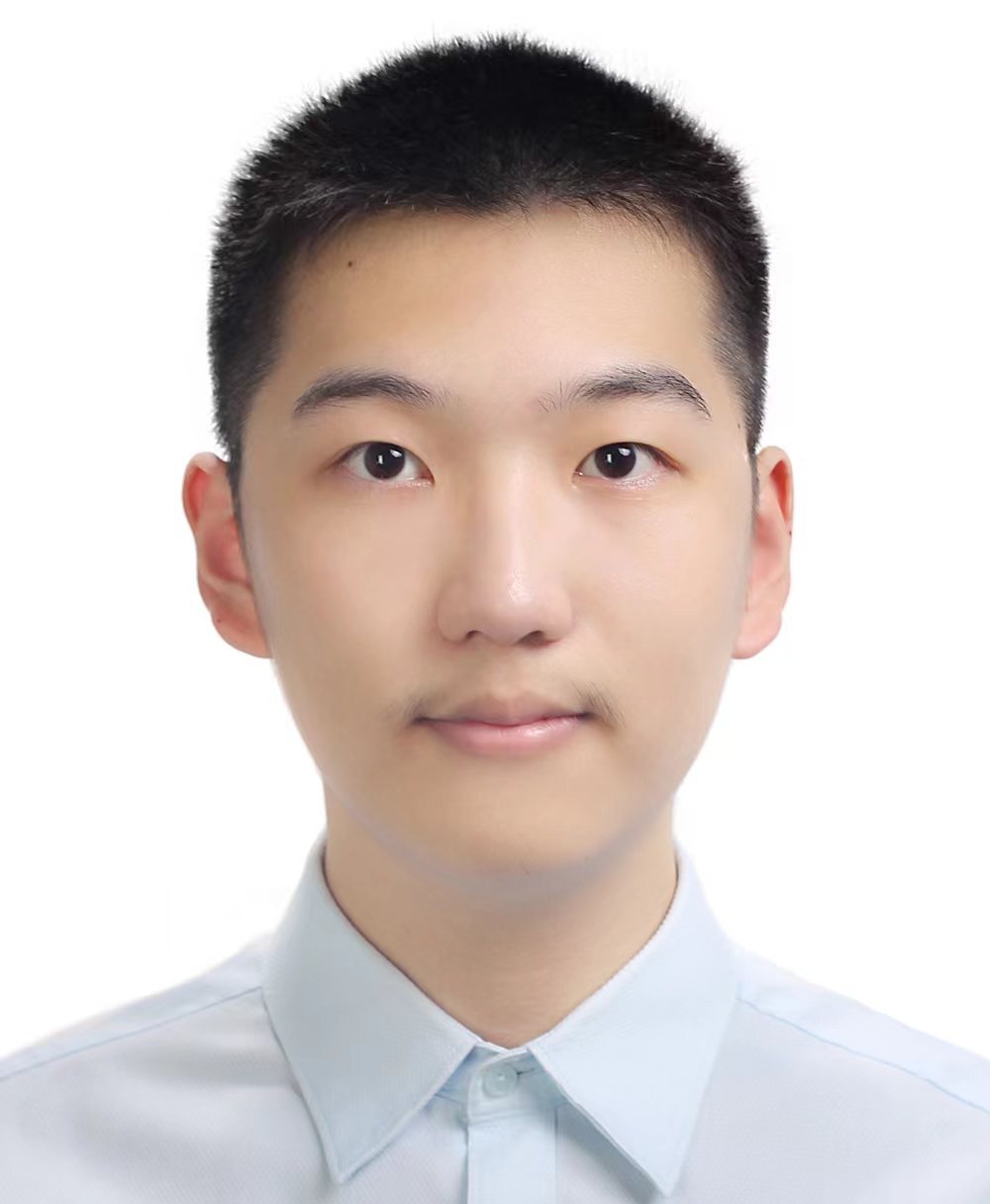}}]{Zhangyu Wang} is currently a rising Senior in Avon Old Farms School, CT, USA. His research interests mainly focus on computer vision and artificial intelligence. 

\end{IEEEbiography}

\begin{IEEEbiography}[{\includegraphics[width=1in,height=1.25in,clip,keepaspectratio]{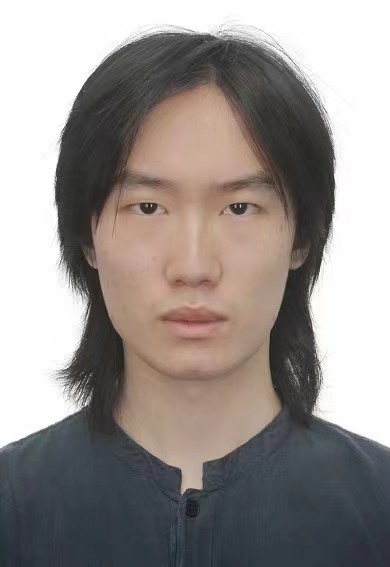}}]{Ningyuan Sun} is currently a student in University of Wisconsin-Madison, WI, USA, pursuing the bachelor degree in Computer Science. His research interest mainly focuses on computer vision and semantic segmentation. 
\end{IEEEbiography}

\EOD

\end{document}